\documentclass{article}

\usepackage{microtype}
\usepackage{graphicx}
\usepackage{subfigure}
\usepackage{booktabs} 
\usepackage{pifont}
\usepackage{xspace}
\usepackage{xcolor}
\usepackage{subcaption}

\usepackage{hyperref}

\usepackage[accepted]{icml2025}

\usepackage{amsmath}
\usepackage{amssymb}
\usepackage{mathtools}
\usepackage{amsthm}
\usepackage[british,UKenglish,USenglish,english,american]{babel}

\usepackage[capitalize,noabbrev]{cleveref}

\theoremstyle{plain}

\theoremstyle{definition}

\theoremstyle{remark}

\usepackage[textsize=tiny]{todonotes}

\newcommand{\ours}{Hibiki\xspace}
\newcommand{\ensuremathmode}[1]{%
  \ifmmode
    #1
  \else
    $#1$
  \fi
}
\newcommand{\softmax}{\mathrm{softmax}}
\DeclareMathOperator*{\argmax}{arg\,max}
\newcommand{\proba}[1]{\mathbb{P}\left[#1\right]}

\icmltitlerunning{High-Fidelity Simultaneous Speech-To-Speech Translation}

\begin{document}

\twocolumn[
\icmltitle{High-Fidelity Simultaneous Speech-To-Speech Translation}



\icmlsetsymbol{equal}{*}

\begin{icmlauthorlist}
\icmlauthor{Tom Labiausse}{kyutai}
\icmlauthor{Laurent Mazar\'e}{kyutai}
\icmlauthor{Edouard Grave}{kyutai}
\icmlauthor{Patrick P\'erez}{kyutai}
\icmlauthor{Alexandre D\'efossez}{kyutai}
\icmlauthor{Neil Zeghidour}{kyutai}
\end{icmlauthorlist}

\icmlaffiliation{kyutai}{Kyutai, Paris, France}

\icmlcorrespondingauthor{\\Hibiki}{hibiki@kyutai.org}

\icmlkeywords{Machine Learning, ICML}

\vskip 0.3in
]



\printAffiliationsAndNotice{} 

\begin{abstract}
\looseness=-1
We introduce \ours, a decoder-only model for simultaneous speech translation. \ours leverages a multistream language model to synchronously process source and target speech, and jointly produces text and audio tokens to perform speech-to-text and speech-to-speech translation.
We furthermore address the fundamental challenge of \textit{simultaneous} interpretation, which unlike its \textit{consecutive} counterpart---where one waits for the end of the source utterance to start translating--- adapts its flow to accumulate just enough context to produce a correct translation in real-time, chunk by chunk.
To do so, we introduce a weakly-supervised method that leverages the perplexity of an off-the-shelf text translation system to identify optimal delays on a per-word basis and create aligned synthetic data. After supervised training, \ours{} performs adaptive, simultaneous speech translation with vanilla temperature sampling. On a French-English simultaneous speech translation task, \ours demonstrates state-of-the-art performance in translation quality, speaker fidelity and naturalness. Moreover, the simplicity of its inference process makes it compatible with batched translation and even real-time on-device deployment. We provide examples\footnote{\href{https://hf.co/spaces/kyutai/hibiki-samples}{https://hf.co/spaces/kyutai/hibiki-samples}} as well as models and inference code.\footnote{\href{https://github.com/kyutai-labs/hibiki}{https://github.com/kyutai-labs/hibiki}}
\end{abstract}

\section{Introduction}
\label{introduction}

\looseness=-1
We introduce \ours{} (``echo'' in Japanese), a system for streaming and expressive speech-to-speech (S2ST) and speech-to-text (S2TT) translation. 
Most work in speech translation has focused on the offline (or \textit{consecutive}) setting where the model has access to the full source utterance before it translates, as it provides useful context while fitting many use cases such as offline video dubbing. A more challenging setting is that of \textit{simultaneous} translation, where translated speech is produced on-the-fly. This task, typically performed by human interpreters, requires a real-time decision-making process to evaluate whether enough context has been accumulated to reliably translate another chunk of speech. When cast as a machine learning problem, this endeavor presents additional challenges such as the lack of speech data aligned at a chunk-level.

\looseness=-1
\ours is a decoder-only model which synchronously receives source speech and generates translated speech by leveraging a multistream architecture. In this context, nested global and local Transformers~\cite{attentionvaswani} jointly model two audio streams by predicting a hierarchy of text and audio tokens for each of them. At inference time, temperature sampling combined with a causal audio codec allows for streaming inputs and outputs. While this architecture was originally introduced by~\citet{moshi} for full-duplex spoken dialogue, we show how it provides a simple and convenient architecture for simultaneous speech translation. To train \ours, we generate synthetic parallel data by translating and resynthesizing the transcript of single-language audio. While this provides pairs of inputs and outputs aligned at the sequence level, this does not allow learning fine-grained alignments. We thus introduce ``contextual alignment'', a simple method based on the perplexity of an off-the-shelf machine translation system~\cite{madlad} to derive word-level alignments. By then introducing proper silences into target speech, we can train \ours to adapt its flow in real-time, without the need for complex inference policies. Moreover, observing that training data varies widely in speaker similarity, we propose to label training examples with categories of speaker similarity, which avoids filtering the training data while allowing to favor high speaker similarity at inference with classifier-free guidance.

\looseness=-1
In a French-English translation task, \ours outperforms previous work in translation quality, speaker similarity and naturalness. We also show how the simplicity of our inference process allows for translating hundreds of sequences in real-time on GPU, and how a distilled model can run in real-time on a smartphone. Human evaluations demonstrate that \ours is the first model to provide an experience of interpretation close to human professionals. We will release our code and models, and a high quality 900 hours synthetic dataset.
\section{Related Work}
\label{related-work}

\begin{figure}[t]
    \centering
    \includegraphics[width=0.9\columnwidth]{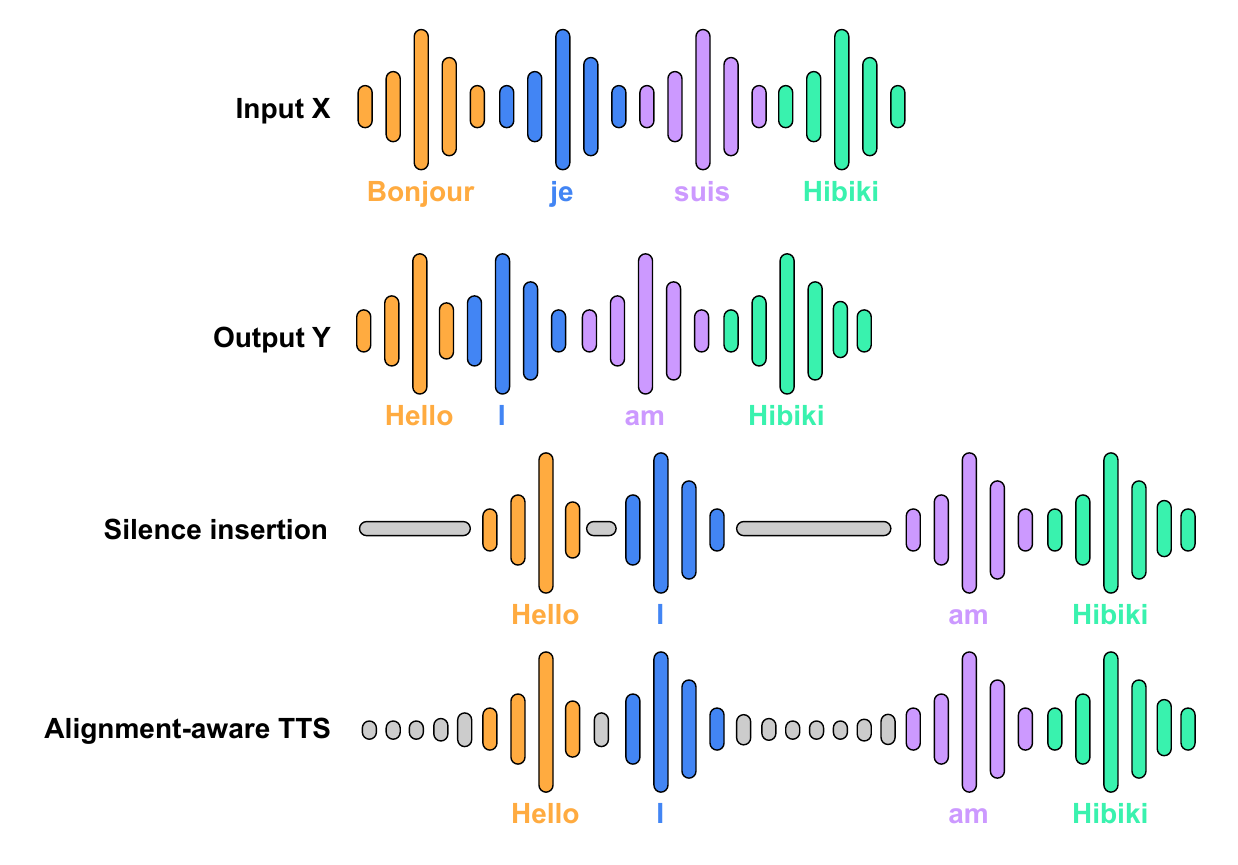}
    \caption{\textbf{Generating aligned interpretations.}
        We extract unsupervised word level contextual alignment, which we lift to audio by either inserting
        silences, or re-synthesising with an alignment aware TTS. See Section~\ref{sec:alignment} for details.
    }
    \label{fig:contextual-delays-waveform}
\end{figure}

\subsection{End-to-end speech translation}
\looseness=-1
Speech translation can be traced back to the early 1990s~\cite{connectionist_translation} with a first generation of systems that combined automatic speech recognition (ASR), machine translation (MT) and text-to-speech synthesis (TTS). While such cascaded approaches allowed for the growth of speech translation~\cite{wahlster2000verbmobil, nakamura2006atr}, they suffer from two main limitations. First, they are subject to compounding errors due to combining separately trained models. This motivated the merging of ASR and MT into a single speech-to-text translation (S2TT) model~\cite{s2tt_berard2016, s2tt_weiss_2017, s2tt_wang_2020, s2tt_wang_2021} that can provide inputs to a TTS model. However, a second limitation of cascaded systems remains: as the input speech goes through a text bottleneck, the non-linguistic information it carries---such as speaker identity or prosody---is lost and cannot be transferred to the output speech. End-to-end speech-to-speech translation (S2ST)~\cite{jia19_translatotron,lee-etal-2022-direct,jia22-translatotron2,audiopalm} addresses this issue by directly predicting target speech from source speech, allowing for retaining paralinguistic information, including voice identity. A notable aspect of most end-to-end S2ST models is that they leverage auxiliary text or phoneme translation tasks in training, that are then discarded~\cite{jia22-translatotron2} or run in parallel~\cite{streamspeech} to the main speech translation task at inference. \ours performs end-to-end S2ST with voice transfer along with S2TT, but instead of running these tasks in parallel, \ours uses the predicted text as a scaffolding for speech generation at inference time. Moreover, since \ours predicts aligned speech and text tokens, it provides word-level timestamps in the target language. 
\subsection{Simultaneous speech translation}
\looseness=-1
While the first attempts at simultaneous speech translation focused on speech-to-text~\citep{ren-etal-2020-simulspeech, ma_streaming_21, real_trans_21}, Seamless~\cite{seamless} and StreamSpeech~\cite{streamspeech} have introduced end-to-end simultaneous S2ST~\cite{real_trans_21,seamless,streamspeech}. Both systems predict discrete speech units with autoregressive models before decoding them to audio using a neural vocoder, and rely on a specific policy for inference. While StreamSpeech translates into a canonical voice, Seamless performs voice transfer from source to target. \ours also performs simultaneous S2ST and S2TT, while transferring voice characteristics. However, \ours relies on a decoder-only model which operates at a constant frame rate and performs inference with simple temperature sampling or greedy decoding. In particular, this allows for batching, unlike StreamSpeech's and Seamless's policies that involve a complex control flow that cannot be batched.
This makes \ours able to translate hundreds of sequences on a single GPU, provides
convenient support for classifier-free guidance, see e.g. Section~\ref{sec:voice_transfer}, and allows to run in real time on device, as shown in Section~\ref{sec:inference_capabilities}.
Human evaluations in Section~\ref{sec:eval_human} show that \ours significantly outperforms Seamless in terms of naturalness and audio quality, getting close to human interpretation.

\section{Method}
\label{method}

\begin{figure}[t]
    \centering
    \includegraphics[width=\columnwidth]{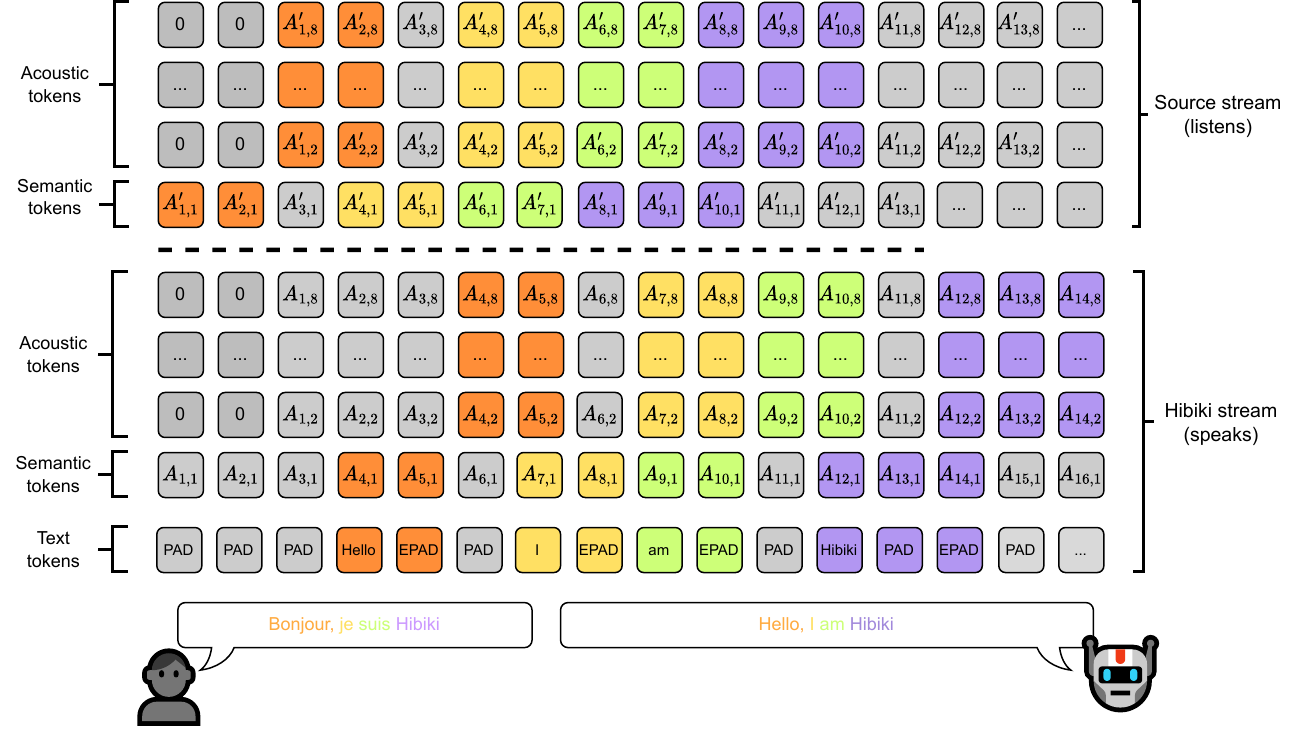}
    \caption{\textbf{Joint sequence modeling with contextual alignment.} From the source stream, \ours predicts its Inner Monologue text
    stream, and audio tokens.
    Its output is aligned for causality, as depicted in Figure~\ref{fig:contextual-delays-waveform}. Figure adapted from ~\citet{moshi}.
    }
    \label{fig:joint-modeling}
\end{figure}
We consider an utterance in a source language represented as
a monophonic waveform $X \in\mathbb{R}^{f_s \cdot d}$, sampled at a frame rate $f_s = 24\,\mathrm{kHz}$, of duration $d$.
Similarly, its translation is given in a target language,
denoted $Y \in \mathbb{R}^{f_s \cdot d}$. We assume $X$ is padded to ensure both have the same duration. Our objective is to model $\proba{Y | X}$.
We further add the constraint that the modeling of $Y$ knowing $X$ 
should be causal and of minimal delay with respect
to the source utterance, e.g. the same constraints that are imposed on a human interpreter in the context of live translation. 
To learn this constraint \textit{via} supervised learning, $Y$ must itself be built to respect this causality constraint.
We first assume that $Y$ respects this constraint, and we present how to model
its distribution. Then, we introduce an information theory criterion 
to verify whether $Y$ is causal with respect to $X$, and to adapt
a non-causal interpretation into a causal one.

\subsection{Modeling}

We build on the framework introduced by \citet{moshi} for the joint modeling of
multiple sequences of discrete tokens, obtained from a neural audio codec.

\subsubsection{Neural audio codec}

We use the pre-trained causal and streaming Mimi codec~\citep{moshi} to encode $X$ and $Y$ into low framerate sequences of discrete tokens.
Mimi consists of an encoder and decoder from and to the waveform domain, and of an information bottleneck using Residual Vector Quantization (RVQ)~\cite{soundstream}. The encoder transforms an input waveform of duration $d$ into a latent vector $U \in \mathbb{R}^{C \times f_r \cdot d}$,
with $C$ the dimension of the latent space, and $f_r = 12.5\,\mathrm{Hz}$ the frame rate. 
$U$ is then projected to its nearest neighbor in a \emph{codebook} table with $N_A$ entries. The residual of the projection is further projected into a second
table with the same cardinality, and so forth until $Q$ projections have been performed. The last residual is discarded, and the decoder is trained to reconstruct the input waveform from the sum of the projected tensors.
The codebooks are trained through exponential moving average, along with a commitment loss~\citep{vqvae}. The rest of the model is trained only through an adversarial loss with feature matching~\citep{moshi}.

For language modeling, we are not interested in the quantized latent vector and its residuals, but in the discrete
indices of the entry in the codebooks it is projected to. We denote those $(A_{t, q}) \in \{1, \ldots, N_A\}^{f_r \cdot d \times Q}$.
For Mimi we have $f_r = 12.5 \,\text{Hz}$ and $Q$ varies up to 32, but we use at most 16. 
Following~\citet{zhang2024speechtokenizer,moshi}, the output
of the first quantization level is trained to replicate semantic information
obtained from a WavLM self-supervised audio model~\citep{wavlm}.
Following conventions of~\citet{audiolm}, we refer to $A_{t, 1}$ as the \emph{semantic} tokens, and $A_{t, q \geq 2}$ as the \emph{acoustic} tokens. The acoustic tokens are arranged in a coarse to fine manner, the first ones have the most importance, and the latest model fine details of the audio and ensuring a smooth perception.

\subsubsection{Joint modeling of discrete audio tokens}

The discrete tokens for audio streams cannot easily be summarized into a single discrete sequence with reasonable cardinality and framerate~\citep{musicgen}. Following \citet{uniaudio,moshi}, we leverage a RQ-Transformer~\citep{rq-transformer} 
to model $(A_{t, q})$ both over the time $t$ and quantizer $q$ axes.
It consists in a large \emph{Temporal} Transformer~\citep{attentionvaswani}, operating at the same framerate $f_r$ as the codec, and being fed all the tokens generated so far, e.g.
for all $t \leq f_r \cdot d$, 
 \begin{equation}
 \label{eq:temp_transformer}
Z_t = \mathrm{Temp}(A_0, \ldots, A_{t - 1}) \in \mathbb{R}^{D}.
\end{equation}
$A_0$ is defined as a deterministic token indicating the start of the generation.
Then, a smaller scale \emph{Depth} Transformer models auto-regressively
the tokens $A_{t, 1}, \ldots, A_{t, Q}$ over the quantizer axis, e.g. for all $t \leq f_r \cdot d$ and $q \leq Q$,
\begin{equation}
\label{eq:dep_transformer}
    l_{t, q} = \mathrm{Dep}(Z_t, A_{t, 0}, \ldots, A_{t, q - 1}) \in \mathbb{R}^{N_a},
\end{equation}
with $A_{t, 0}$ also a special token, and with the goal of having,
\begin{equation*}
    \softmax(l_{t, q}) \approx \proba{A_{t, q} | A_{0}, \ldots, A_{t-1}, A_{t, 0}, \ldots A_{t, q - 1}}
\end{equation*}

Following \citep{musicgen,moshi}, we further introduce an acoustic delay of 2 time steps, meaning that we model $(\tau(A)_{t, q})$ instead of $A$,
\begin{equation}
\begin{cases}
\begin{array}{lll}
    \tau(A)_{t, 1} &= A_{t, 1} &\quad\forall\, t \\
    \tau(A)_{t, q} &= A_{t - 2, q} &\quad\forall\, t \geq 3, \forall\, q \geq 2\\
    \tau(A)_{t, q} &= 0 &\quad\forall\, t < 3, \forall\, q \geq 2, \\
\end{array}
\end{cases}
\end{equation}
with 0 being a special token. The delay is removed before decoding audio with
the codec.

\subsubsection{Translation as multistream modeling}
\label{sec:multistream}

We have presented how the RQ-Transformer given by eq. \eqref{eq:temp_transformer} and \eqref{eq:dep_transformer} allows
for jointly modeling multiple discrete streams of tokens.
We adapt this framework for the task of joint speech-to-speech
and speech-to-text simultaneous translation.
We concatenate the audio tokens
$A^Y$ obtained from the target interpretation $Y$, with
the tokens $A^X$ from the source utterance $X$ along the $q$-axis, e.g. 
\begin{equation}
    \bar{A} = \mathrm{concat_q}\left[\tau(A^Y), \tau(A^X)\right].
\end{equation}
We observe a benefit from modeling the tokens $A^X$ at train time,
although at inference time, predictions for those tokens are skipped
and actual tokens for the input are used instead.

\citet{moshi} showed generating an Inner Monologue, i.e. padded text tokens aligned with the content of the generated audio, is beneficial to the quality
and stability of the generated audio. This is similar to multi-task learning where the translation is predicted both in the audio
and text domain. \ours thus also predicts a text stream corresponding
to the transcription of the output $Y$, with sufficient padding between words to keep them aligned with the audio. Note that unlike previous multi-task
translation work, \ours makes active use of this capability at inference time.
We denote $W_t$ the text stream, with cardinality $N_W$ and the same frame rate $f_r$
as the audio streams.

\subsubsection{Architectural details}
\label{sec:arch_details}

We provide architectural hyper-parameters in Section~\ref{sec:arch_hyperparams}.
At time-step $t$, the tokens from the step $t-1$, e.g. $\tau(A^X)_{t-1}$,
$\tau(A^Y)_{t-1}$, and $W_{t-1}$, are fed into dedicated embedding tables, whose contributions are summed.
For the first time step $t = 1$, a BOS token is used instead.
We then use standard Transformer layers~\citep{attentionvaswani}, with gated SiLU activation~\citep{shazeer2020glu,hendrycks2016gaussian}.
A linear layer maps its output $Z_t$ to logits for the text token $W_t$.
The Depth Transformer then operates for $2 \cdot Q$ steps:
the first half to estimate the logits for the output stream,
and the next half for the input stream.
Each depth step $q$ takes as input $Z_t$ summed with
a learnt embedding of the previous audio token $\bar{A}_{t, q - 1}$, or $W_t$ for $q = 1$.

\subsection{Alignment and synthetic interpretation data}
\label{sec:alignment}

\begin{figure}[t]
    \centering
    \includegraphics[width=1.0\columnwidth]{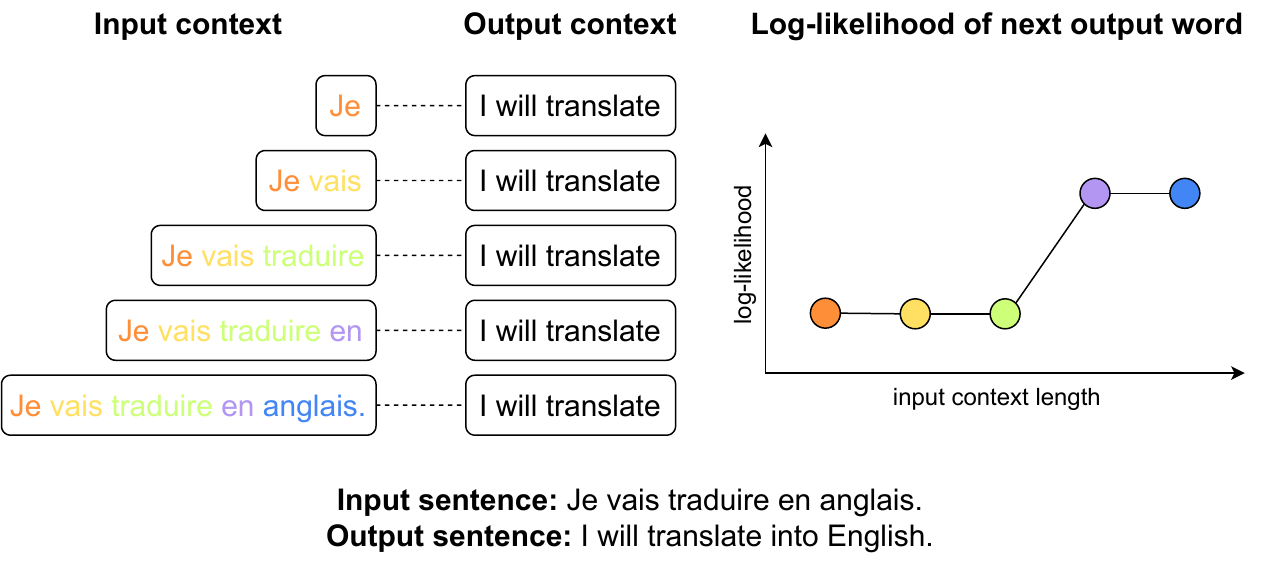}
    \caption{\textbf{Contextual alignment.}
    We compute the log-likelihood of the word ``into'' with a pre-trained text translation model, for various input truncations. Once the matching source word "en" appears, we observe a large increase in log-likelihood,  see eq.~\eqref{eq:ctx_align}.}
    \label{fig:contextual-delays-tokens}
\end{figure}

\looseness=-1
We have assumed pairs $(X, Y)$
that respect the constraint of simultaneous interpretation.
We now introduce an unsupervised criterion to estimate and enforce causality
dependencies between the source and target utterances.

\subsubsection{Text domain alignments}
\label{sec:text-domain-align}
Let us first express formally those constrained in the text domain.
Let us take $S = (S_1, \ldots, S_n)$ the sequence of words
in the utterance $X$, and $T = (T_1, \ldots, T_m)$
that in $Y$.

\paragraph{Ideal alignment.}
We seek to define an ideal alignment $(a^{\text{ideal}}_{j}) \in \{1,\ldots, n\}^m$, where
$a^{\text{ideal}}_{j}$ indicates the index of the word in $S$ that the $j$-th word in $T$
should wait for to minimize the uncertainty on $T_j$. 
Any alignment strictly less conservative than $a^{\text{ideal}}$
would risk the model hallucinating at inference if trained on.
Any alignment strictly more conservative would still be causal,
but introduce more latency.

\paragraph{Contextual alignment.}
We introduce a criterion to estimate $a^{\text{ideal}}$.
Let's denote the conditional log-likelihood
\begin{equation}
    \label{eq:delay_proba_gt}
    \log(p_{j, i}) = \log\left(\proba{T_j | S_1, \ldots S_{i}, T_{1}, \ldots, T_{j-1} }\right),
\end{equation}
we expect $\log p_{j, i}$ to increase with $i$,
as more context is beneficial.
We conjecture that $\delta_{j,i} = \log(p_{j, i}) - \log(p_{j, i - 1})$ is maximal for $i = a_{j}$.
We compute an estimate $\log(\hat{p}_{j, i})$ of 
$\log(p_{j, i})$ with an off-the-shelf 
text translation language model MADLAD-3B~\citep{madlad}, by feeding it
truncated input up to the $i$-th word, which we use to define a \emph{contextual} alignment, illustrated in Figure~\ref{fig:contextual-delays-tokens},
\begin{equation}
\label{eq:ctx_align}
    a^{\text{ctx}}_j = \argmax_{i \leq n}  \left[\log(\hat{p}_{j, i}) - \log(\hat{p}_{j, i - 1}) \right].
\end{equation}
Examples of alignments are given in the Appendix, Figure~\ref{fig:examples-contextual-alignments}.

\subsubsection{Audio domain alignments}
\label{sec:audio-domain-align}

Given $(X, Y)$, we transcribe both with timestamps with 
a Whisper model~\citep{whisper,lintoai2023whispertimestamped}
and apply eq. \eqref{eq:ctx_align}.
The pair $(X, Y)$ respects
the alignment $(a^{\textrm{ctx}}_j)$ if the timestamp of the $j$-th word in $Y$
comes after the timestamp of the $a_j$-th word in $X$. To reduce
the impact of errors, we require $Y$
to lag by at least 2 seconds compared to the contextual alignment, and eliminate spikes higher than 25\% of the average delay over
a window of 5 words.

\paragraph{Silence insertion.}
If $Y$ doesn't respect the alignment,
one can simply transform it by inserting sufficient silences
before a word, as illustrated in Figure~\ref{fig:contextual-delays-waveform}.
This comes, however, with two limitations: \\(i) silence insertion
can lead to hard cuts when the timestamps are inaccurate
or no pause exists between words;\\ (ii) the corrected $Y$ might be arbitrarily
late on the ideal alignment, e.g. if the speech rate is slower in $Y$ than in $X$.\\
We apply this method during the speech translation training.

\paragraph{Alignment-aware TTS.}
We obtain more natural aligned data by (re)-synthesising $Y$
with a TTS model able to follow hard and soft constraints on word locations,
along with accurate speaker conditioning.
For existing datasets, this can have the added benefit of improving the word error rate and the speaker similarity, as illustrated in Section~\ref{sec:voice_transfer}.
Following~\citep{moshi}, Appendix C, we train a TTS with both audio and a synced text stream as output, along with voice conditioning. The text stream is constrained to match
exactly text to generate, with the model having only the freedom to insert padding tokens. The audio output is late on the text, so that its content is
conditioned by it, both for content and timestamps. 
If the TTS is early on the alignment $a^{\textrm{ctx}}$, 
padding tokens are forced to delay the next word.
When the TTS is lagging on its target, a penalty
is added on the logits of the padding token. The penalty scales from 0 to -2 as
the lag increases from 1 to 2 seconds. This increases smoothly the rate of 
speech to catch up with the source audio.
We perform 6 to 8 generations per input, and select the best one based on
word error rate first, and speaker similarity second.
We apply this only for the fine-tuning speech translation dataset.

\begin{figure}[t]
        \centering
        \includegraphics[width=0.9\columnwidth]{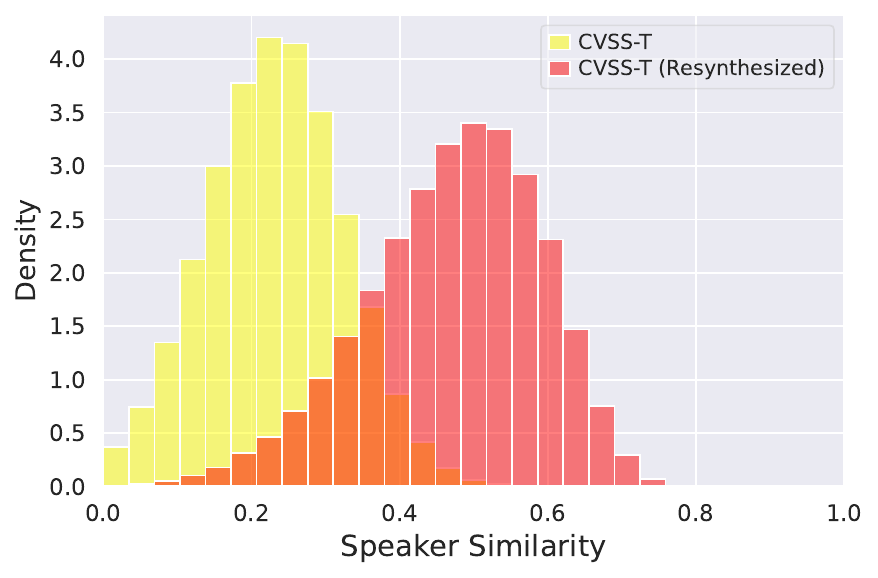}
    \caption{Speaker similarity between source and target speech in CVSS-T training data, before and after resynthesis.
    }
    \label{fig:speaker-sim-histogram}
\end{figure}

\subsection{Voice Transfer}
\label{sec:voice_transfer}
\paragraph{Improving voice transfer data.}
\looseness=-1
Training speech translation models with voice transfer typically amounts to supervised training on synthetic paired sequences of the same speaker.
In particular, CVSS-T~\cite{cvss} is the standard training set for S2ST with voice transfer and provides such artificial targets. However, Figure~\ref{fig:speaker-sim-histogram} shows that the average speaker similarity---as measured by the cosine similarity between speaker embeddings of source and target--- on this dataset is very low with an average of $0.23$. As a calibration, state-of-the-art cross-lingual voice transfer lies around $0.40$~\cite{audiopalm}. We thus also regenerate CVSS-T with our alignment-aware TTS, as it allows for voice transfer. As shown in Figure~\ref{fig:speaker-sim-histogram}, the resynthesized CVSS-T displays a higher similarity, with an average of $0.47$. Yet, our training mixture which combines synthetic data and resynthesized CVSS-T still covers a wide range, with a significant mass below $0.40$.

\paragraph{Conditional training.} Filtering training data to only keep pairs of examples with a high similarity would improve voice transfer, but the resulting reduction in training data would likely affect translation quality (e.g. keeping only samples with a speaker similarity above $0.40$ would remove ~45\% of training data). We rather rely on conditional training~\cite{socher_conditioned_2019, goal_conditioned_bowman} to inform the generative model of how reliable each training example is in terms of voice transfer. We label each training sample with a discrete ``voice transfer score'' in $\{\texttt{very\_bad},~\texttt{bad},~\texttt{neutral},~\texttt{good},~\texttt{very\_good}\}$ based on quantiles of speaker similarity, each label being associated to a learnable embedding added to the model's inputs at every timestep. Importantly, the quantiles are computed \emph{before} combining the synthetic data and CVSS-T to guarantee that the model does not associate a specific label to a specific dataset rather than the actual speaker similarity. At inference time, we always pass the $~\texttt{very\_good}$ label.

\looseness=-1
\paragraph{Classifier-free guidance.} Following~\citet{audiogen} we can increase the impact of the conditioning by using classifier-free guidance. We compute logits both with conditionings $~\texttt{very\_good}$ and $~\texttt{very\_bad}$ and sample from:
\begin{equation}
    \gamma l_{t, q}^{~\texttt{very\_good}} + (1 - \gamma) l_{t, q}^{~\texttt{very\_bad}}, 
\end{equation}
which is compatible with real-time inference by producing both sets of logits with a batch size of 2. Section~\ref{sec:results} shows that it significantly improves voice transfer.

\section{Experiments}
\label{sec:experiments}
\subsection{Architectural hyper-parameters}
\label{sec:arch_hyperparams}

\looseness=-1
\ours consists of a Temporal Transformer
with a latent dimension of 2560 (7040 for the SiLU gating), 24 layers, 20 heads and local attention over 500 tokens, i.e. 2.2B parameters and a 40s context. The Depth Transformer initially follows
~\citet{moshi}, i.e. 6 layers per codebook,
a latent dimension of 1024 (2816 for the gating), and 64 dimensions per head. It models $Q=16$ audio codebooks
for the output stream, and the same for the input stream (only at training). To reduce the footprint of the Depth Transformer we distill it post-training
into a smaller one, with 4 layers per codebook, weight sharing for codebooks 9 to 16, low-rank embedding tables of dimension 128, and a dimension of 2048 for the gating. This reduces its size from 1.1B parameters to 449M parameters, for a total of 2.7B parameters.
We also train \ours-M as a distilled version of \ours, a 1.7B variant capable of running real-time on device, with a Temporal Transformer with
a latent space of dimension 2048, and 16 layers, and only 8 codebooks levels per stream.

\subsection{Training protocol}

We train a French-English speech translation system through the following steps, each with a cosine learning rate schedule and AdamW~\citep{adamw}, with a weight decay of 0.1, and momentum parameters of (0.9, 0.95).

\textbf{Text pretraining.}
We first pretrain the \textit{Temporal} Transformer from scratch on multilingual text-only data using next token prediction, for 600K steps, with a batch of 1,024 sequences of length 4,096. We use a cosine learning rate schedule, with 2K warmup steps and a maximum value of $4.8\cdot10^{-4}$. Our training dataset is made of filtered web pages from Common Crawl, as well as curated sources such as Wikipedia, StackExchange or scientific articles and it contains 12.5\% of multilingual documents.

\textbf{Audio pretraining.}
Starting from the pretrained text model, we perform an audio pretraining on non-parallel French and English data with a single stream as done by \citet{moshi}. We train for 1,450K steps with a batch size of 144 and a learning rate of $2\cdot10^{-4}$. Then we duplicate the weights of the Depth Transformer for multistream modeling.

\textbf{Speech translation training.}
We build a French-English speech translation dataset of approximately 40K hours in each language. Starting from a collection of expressive audio content in French, we extract roughly 2.3M single speaker utterances each with a duration around 60 seconds. We transcribe these segments with Whisper \citep{whisper}, using the large-v3 model. We rely on PySBD \citep{pysbd} to segment each transcript into sentences and use MADLAD-3B \citep{madlad} to translate them individually, before joining them back into a translated English transcript. We synthesize each with the TTS
described in Section~\ref{sec:audio-domain-align}, conditioned on the original French speaker identity with a 10s utterance. 
We apply the silence insertion technique described in Section~\ref{sec:audio-domain-align} to obtain simultaneous interpretation pairs. We train for 150K steps with a batch size of 96, a learning rate of $3\cdot10^{-5}$ and compute the loss on both the source and the target streams. We use conditional training on speaker similarity as explained in \ref{sec:voice_transfer} and apply noise augmentation techniques on the source audio. For each training pair, we introduce a special input EOS token on all audio tokens of the source at the first frame after the end of its speech utterance and use another special EOS token on the text tokens stream to indicate the end of the model speech utterance. 

\textbf{Speech translation fine-tuning.}
We use alignment-aware TTS generations introduced in Section~\ref{sec:audio-domain-align} to build a synthetic dataset composed of long-form utterances and an improved version of CVSS-T/train, with natural pauses and high speaker similarity, totaling close to 900 hours. 
We fine-tune for 8K steps with a batch size of 8, a learning rate of $2\cdot10^{-6}$, conditional training on the speaker similarity, special EOS tokens, and apply the loss to both streams.

\textbf{Training of \ours-M.}
It goes through the same text and audio pre-training stages. During 
the speech translation training it is soft distilled from \ours, before
going through the same fine-tuning step (without distillation).

\begin{table}[t]
\caption{\textbf{Comparison with offline baselines.} We also report performance from a closed-source streaming model (*) as it uses the same evaluation protocol.}
\label{tab:offline}
\vskip 0.15in
\begin{center}
\begin{sc}
\resizebox{\linewidth}{!}{%
\begin{tabular}{lc}
\toprule
Model & ASR-BLEU \\
\midrule
Translatotron \cite{jia19_translatotron} & 17.0 \\
Translatotron 2 \cite{jia22-translatotron2} & 26.0 \\
S2UT~\cite{lee-etal-2022-direct} & 22.2 \\
UnitY \cite{inaguma-etal-2023-unity} & 27.8 \\
DASpeech \cite{fang2023daspeech} & 25.0 \\
RNN-Transducer*~\cite{rnn-t-s2st} & 25.4 \\
StreamSpeech (Offline)~\cite{streamspeech} & 28.5 \\
\midrule
\ours & \textbf{30.5} \\

\bottomrule
\end{tabular}}
\end{sc}
\end{center}
\vskip -0.1in
\end{table}

\begin{table*}[t]
\caption{Objective comparison of Hibiki with StreamSpeech~\citep{streamspeech} and Seamless~\citep{seamless}.}
\label{tab:streaming_big_table}
\vskip 0.15in
\begin{center}
\begin{sc}
\resizebox{0.875 \linewidth}{!}{%
\begin{tabular}{lcccccccccc}
\toprule
 & \multicolumn{5}{c}{Short-form (CVSS-C Fr-En test)} & \multicolumn{5}{c}{Long-form (Audio-NTREX)} \\
\cmidrule(lr){2-6}\cmidrule(lr){7-11}
      &                   & ASR               & Speaker           & End & & & ASR & Speaker & End & \\
 Model & BLEU ($\uparrow$) & BLEU ($\uparrow$) & Sim. ($\uparrow$) & Offset ($\downarrow$) & LAAL ($\downarrow$) & BLEU ($\uparrow$) & BLEU ($\uparrow$) & Sim. ($\uparrow$) & Offset ($\downarrow$)  & LAAL ($\downarrow$) \\   
\midrule
StreamSpeech
 & 26.4 & 25.4 & - & 1.6 & 2.8
 & 0.1 & 0.1 & - & N/A & N/A \\
Seamless
 & 37.0 & 33.8 & 0.30 & \textbf{1.4} & \textbf{2.8}
 & 25.4 & 23.9 & 0.43 & \textbf{1.6} & \textbf{4.2} \\
 \midrule
 \ours-M & 37.5  & 33.7 & 0.34 & 2.8 & 3.5
 & 25.9 & 25.0 & 0.39 & 2.3 & 5.5 \\
\ours
 & \textbf{39.2} & \textbf{35.5} & \textbf{0.41} & 2.9 & 3.4
 & \textbf{27.5} & \textbf{26.6} & \textbf{0.48} & 2.7 & 5.0 \\
\bottomrule
\end{tabular}}
\end{sc}
\end{center}
\vskip -0.1in
\end{table*}

\subsection{Evaluation metrics and baselines}
\label{sec:eval-metrics}

\looseness=-1
We first compare \ours to several offline baselines as well as a closed source streaming baseline~\cite{rnn-t-s2st}.
We then perform a comparison between \ours and the two existing methods for simultaneous translation: Seamless~\cite{seamless} and StreamSpeech~\cite{streamspeech} with a chunk size of 2560ms.

\textbf{Translation quality.} We evaluate translation quality by transcribing generated speech and computing a BLEU score~\cite{sacrebleu} with the reference, referred to as ASR-BLEU. When comparing to offline baselines in Table~\ref{tab:offline}, we replicate the setting of \citet{streamspeech} for a fair comparison. However, we observe that this ASR model makes frequent mistakes, so in subsequent experiments we rather transcribe using Whisper medium
~\cite{whisper} and evaluate the BLEU score between ground-truth and hypothesis after normalization.\footnote{\href{https://github.com/openai/whisper/blob/main/whisper/normalizers/english.py}{github.com/openai/whisper/blob/main/whisper/normalizers}} Since Seamless, StreamSpeech and \ours also produce a text translation, we also evaluate their BLEU score using the same text normalization.

\looseness=-1
\textbf{Audio quality and naturalness.} Human raters evaluate the audio quality of generated speech and its naturalness. We evaluate the latter as a proxy for ``realism'': are the flow and prosody natural, are pauses smooth and properly placed or are there abrupt cuts? We compute each score for each model by averaging Mean Opinion Scores between 1 and 5 across 30 samples, each sample being evaluated by 15 raters.

\vspace{-1em}
\looseness=-1
\paragraph{Cross-lingual speaker similarity.} For objective evaluation, we use a standard model for speaker verification\footnote{\href{https://github.com/microsoft/UniSpeech/tree/main/downstreams/speaker\_verification\#pre-trained-models}{github.com/microsoft/UniSpeech} (``WavLM Large'')} based on WavLM~\cite{wavlm} and report the cosine similarity between the embeddings of the source and the generated speech. To mitigate potential biases due to using the same speaker verification model for conditional training (see Section~\ref{sec:voice_transfer}), we also collect human judgments where raters are asked to rate the similarity to the source audio.

\looseness=-1
\textbf{Latency.} A metric for S2ST latency is the End Offset, which is the time (in seconds) between the end of the last word of the source and that of the last word in the output. 
We also measure the Length-Adaptative Average Lagging (LAAL) following the method described by \citet{laal}: it approximates the average time (in seconds) between the pronunciation of a source word and its translation, without requiring word-level alignments. We rely on word-level emission timestamps $(d_i)_{1 \dots n_{\mathrm{gen}}}$ produced by Whisper for $n_{\mathrm{gen}}$ words in the generated speech. We define $\delta = \frac{\Delta_{\mathrm{source}}}{max(n_{\mathrm{gen}}, n_{\mathrm{ref}})}$ where $\Delta_{\mathrm{source}}$ is the duration of the source speech and $n_{\mathrm{ref}}$ the number of words in the reference translation. The LAAL score is then computed as $\frac{1}{n_{\mathrm{max}}}\sum_{i=1}^{n_{\mathrm{max}}} d_i - (i-1)\delta$ where $n_{max} = \mathrm{min}\{i|d_i \geq \Delta_{\mathrm{source}}\}$.

\subsection{Evaluation datasets}

\textbf{Short-form data.}
We evaluate models on the Fr-En task of CVSS~\cite{cvss}. While it is the standard benchmark for S2ST and allows comparisons with previous models, we observe that 99\% of its sequences are shorter than 10 seconds. We thus extend our evaluation to long-forms.

\textbf{Long-form data.}
We collect long-form speech translations by recording bilingual speakers as they read Fr-En translations from the NTREX~\cite{ntrex} text corpus. This speech corpus, that we name Audio-NTREX contains 10 hours of real human speech in each language with 10 different speakers and an average of 50 sec. per utterance.

\textbf{Real interpretation.}
\looseness=-1
To compare with human interpreters, we use 90 real interpretations of the European Parliament from VoxPopuli~\cite{voxpopuli} where translations contain the source speech at a lower volume. For a fair comparison, we also add the lowered source speech to generations of \ours and Seamless (see our external webpage for samples).

\subsection{Inference configuration}

\looseness=-1
We encode audio with the streaming codec and feed the tokens to \ours while decoding the output tokens to obtain streaming translation. 
At the end of the input, we send the EOS token to our model, and keep sampling until it produces its own EOS. Inference parameters are cross validated independently for each dataset using a held-out 8\% of Audio-NTREX and the valid split of CVSS-C. 
The optimal parameters are $\gamma = 3.0$, a temperature of 0.8, top-k of 250 for audio tokens and 50 for text tokens for Audio-NTREX. On CVSS, the same configuration is used except for text tokens that are sampled with a temperature of 0.1. We conjecture that the lower text temperature typically improves translation but can lead to producing an EOS token too early.

\subsection{Results}
\label{sec:results}
\paragraph{Translation quality.}
Table~\ref{tab:offline} compares \ours with offline baselines that have access to the complete source audio when translating. Despite performing simultaneous translation, \ours outperforms all models, including the offline variant of StreamSpeech. Table~\ref{tab:streaming_big_table} benchmarks \ours against available baselines for simultaneous translation. In the short-form setting, our model outperforms StreamSpeech and Seamless at the cost of an average 0.7s of additional lagging. The long-form dataset represents a more significant challenge, as StreamSpeech does not manage to produce intelligible translations. \ours outperforms Seamless, again with a latency higher by an average of 0.8s.

\begin{table}[t]
\caption{\textbf{Human evaluation.} Raters report Mean Opinion Scores (MOS) between 1 and 5.}
\label{tab:human_eval}
\vskip 0.15in
\begin{center}
\begin{scriptsize}
\begin{sc}
\begin{tabular}{lccc}
\toprule
Model & Quality & Speaker Sim. & Naturalness \\
\midrule
Ground-truth & 4.18 $\pm$ 0.07 & - & 4.12 $\pm$ 0.08 \\
\midrule
Seamless & 2.22 $\pm$ 0.08 & 2.86 $\pm$ 0.12 & 2.18 $\pm$ 0.09 \\
\ours & \textbf{3.78} $\pm$ 0.09 & \textbf{3.43} $\pm$ 0.10 & \textbf{3.73} $\pm$ 0.09 \\
\bottomrule
\end{tabular}
\end{sc}
\end{scriptsize}
\end{center}
\vskip -0.1in
\end{table}

\looseness=-1
\paragraph{Audio fidelity.}
\label{sec:eval_human}
Objective evaluations for speaker similarity, as reported in Table~\ref{tab:streaming_big_table}, show that \ours demonstrates significantly better voice transfer than Seamless (we do not evaluate StreamSpeech as it does not perform voice transfer). Human evaluations reported in Table~\ref{tab:human_eval} confirm this result and furthermore show a much higher quality and naturalness than Seamless, that get close to that of ground-truth audio from professional human interpreters. This means that not only \ours produces high-quality audio, but that it inserts smooth and natural pauses into its flow.

\begin{table}[t]
\caption{Ablations.}
\label{tab:ablation_delay}
\vskip 0.15in
\begin{center}
\begin{scriptsize}
\begin{sc}
\begin{tabular}{lccc}
\toprule
Model & ASR-BLEU & LAAL\\
\midrule
No lag & 4.2 & 2.46 & \\
Constant lag (2s) & 10.0 & 2.49 & \\
Constant lag (10s) & 22.5 & 9.02 & \\
Sentence alignment & 25.6 & 21.49 & \\
\midrule
No Inner Monologue & 17.1 & 14.34 \\
No audio pretraining & 14.6 & 5.12 \\
\midrule
Hibiki & 26.6 & 5.0 & \\
\bottomrule
\end{tabular}
\end{sc}
\end{scriptsize}
\end{center}
\vskip -0.1in
\end{table}

\begin{table}[t]
\caption{Ablations on classifier-free guidance.}
\label{tab:ablation-cfg}
\vskip 0.15in
\begin{center}
\begin{scriptsize}
\begin{sc}
\begin{tabular}{lcc}
\toprule
CFG param. & ASR-BLEU & Speaker Sim.\\
\midrule
No CFG & 26.0 & 0.42 \\
$\gamma = 3.0$ (default) & 26.6 & 0.48 \\
$\gamma = 10.0$ & 18.9 & 0.44 \\
\bottomrule
\end{tabular}
\end{sc}
\end{scriptsize}
\end{center}
\vskip -0.1in
\end{table}

\begin{figure}[t]
    \centering
    \includegraphics[width=1.0\columnwidth]{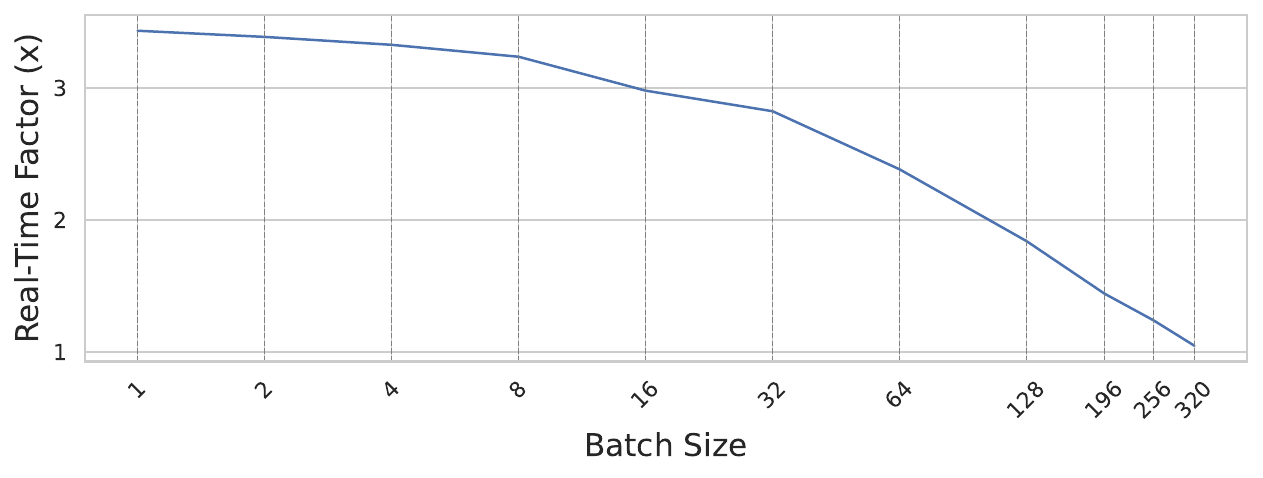}
    \caption{Batched inference speed of Hibiki on a H100 SXM.}
    \label{fig:batched-inference}
\end{figure}

\looseness=-1
\textbf{Ablation: alignment strategies.} We compare our contextual alignment method with alternatives. Table~\ref{tab:ablation_delay} shows that applying no lag to the target speech during training results in very low translation quality, which can be expected since the model lacks context to produce the translation. Adding lag in training examples improves the ASR-BLEU, with 10 seconds representing a reasonable value, however the resulting average latency (as represented by LAAL) is much worse than using a contextual alignment, as the model does not adapt its flow to the context. A middle-ground between a constant lag and contextual alignment is that of ``sentence alignment'' which simply moves the start of each output sentence to the end of the corresponding source sentence. This improves translation quality, however degrading the latency even more. Overall, contextual alignment provides the best trade-off between translation quality and latency.

\textbf{Ablation: Classifier-free guidance.} Table~\ref{tab:ablation-cfg} shows that using the $\texttt{very\_good}$ label provides a speaker similarity of 0.42, similar to that of Seamless (0.43). Using classifier-free guidance with $\gamma = 3.0$ significantly improves it without significantly hurting translation quality, while increasing its weight too much results in degraded performance due to unintelligible speech. Supplementary material interestingly illustrates how increasing $\gamma$ to extreme values results in an exaggerated French accent (the source language in our experiments), which we can attribute to biases in the speaker model used to label our data.

\textbf{Ablation: General ablations.} Section~\ref{sec:multistream} describes how jointly predicting text tokens serves as a scaffolding for audio generation. Table~\ref{tab:ablation_delay} illustrates this claim as training \ours in a unimodal fashion, without predicting text outputs, results much worse performance, as does starting from a pretrained text LM and training directly for S2ST.

\begin{figure}[t]
    \centering
    \includegraphics[width=1.0\columnwidth]{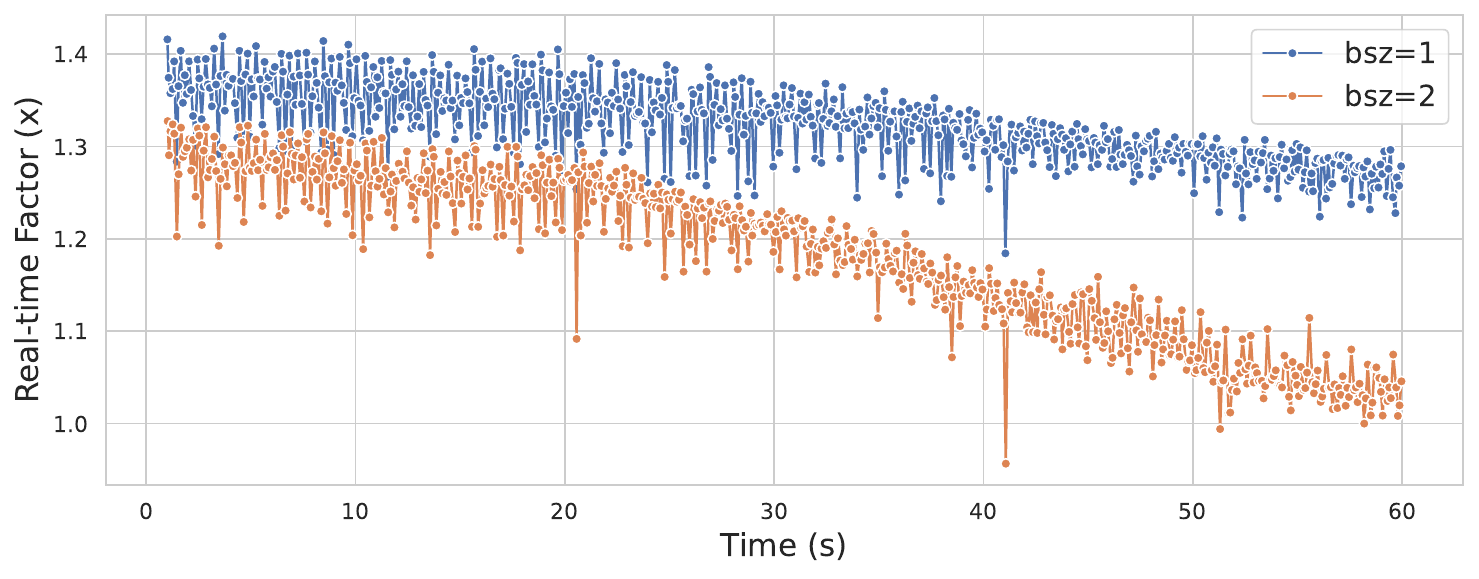}
    \caption{Inference speed of Hibiki-M on an iPhone 16 Pro.}
    \label{fig:iphone-inference}
\end{figure}

\subsubsection{Inference capabilities}
\label{sec:inference_capabilities}
\looseness=-1
\textbf{Batched inference.} \ours's inference uses temperature sampling with a constant framerate. This makes it easy to perform streaming classifier-free guidance as well as batching the processing of several sources of speech at the same time. This is unlike Seamless and StreamSpeech whose complex inference policies are difficult to batch as they require dynamic, irregular decisions for each sequence. Figure~\ref{fig:batched-inference} shows that \ours remains faster than real-time on a H100 even when processing 320 sequences in parallel (or 160 with classifier-free guidance).

\looseness=-1
\textbf{On-device inference.} Our distilled \ours-M is competitive with Seamless on short-form and long-form translation as shown in Table~\ref{tab:streaming_big_table}. We attribute the lower speaker similarity on long-form audio to the lower number of quantizers modeled by \ours-M (8 instead of 16) which results in a twice lower audio bitrate.  Figure~\ref{fig:iphone-inference} shows inference traces of \ours-M on an iPhone 16 Pro. \ours-M remains faster than real-time along a minute of inference, even with a batch size of 2 which is necessary for classifier-free guidance. Training \ours-M with sliding window attention would furthermore improve real-time factor along time.

\subsubsection{Limitations}
This study focuses on a single translation task (French to English) and scaling to more languages could benefit from MADLAD which is massively multilingual, however it would require training TTS systems on more languages. Moreover, while \ours reaches 35.5 ASR-BLEU against CVSS-C ground-truth targets, it reaches 47.9 ASR-BLEU if compared to MADLAD text translations instead. This shows that \ours is excellent at predicting translations that could be produced by MADLAD, and training it to predict pseudo-targets from better or more diverse translations can improve translation quality w.r.t ground-truth targets.
\section{Conclusion}
\label{sec:conclusion}
\looseness=-1
We introduce \ours, a model for simultaneous speech and text translation. \ours leverages a multistream architecture that casts live translation as simple temperature sampling, thanks to a weakly-supervised method for aligning paired data. \ours is competitive with the state-of-the-art in terms of translation quality, while demonstrating a much more natural flow and better voice transfer, close to human interpretation. Moreover, \ours is compatible with streaming batched inference, which facilitates large-scale deployment, while the smaller \ours-M runs in real-time on a smartphone.
\clearpage

\bibliography{references}
\bibliographystyle{icml2025}

\newpage
\appendix
\onecolumn
\section{Appendix}


\begin{figure}[h]
    \centering
    \includegraphics[width=\columnwidth]{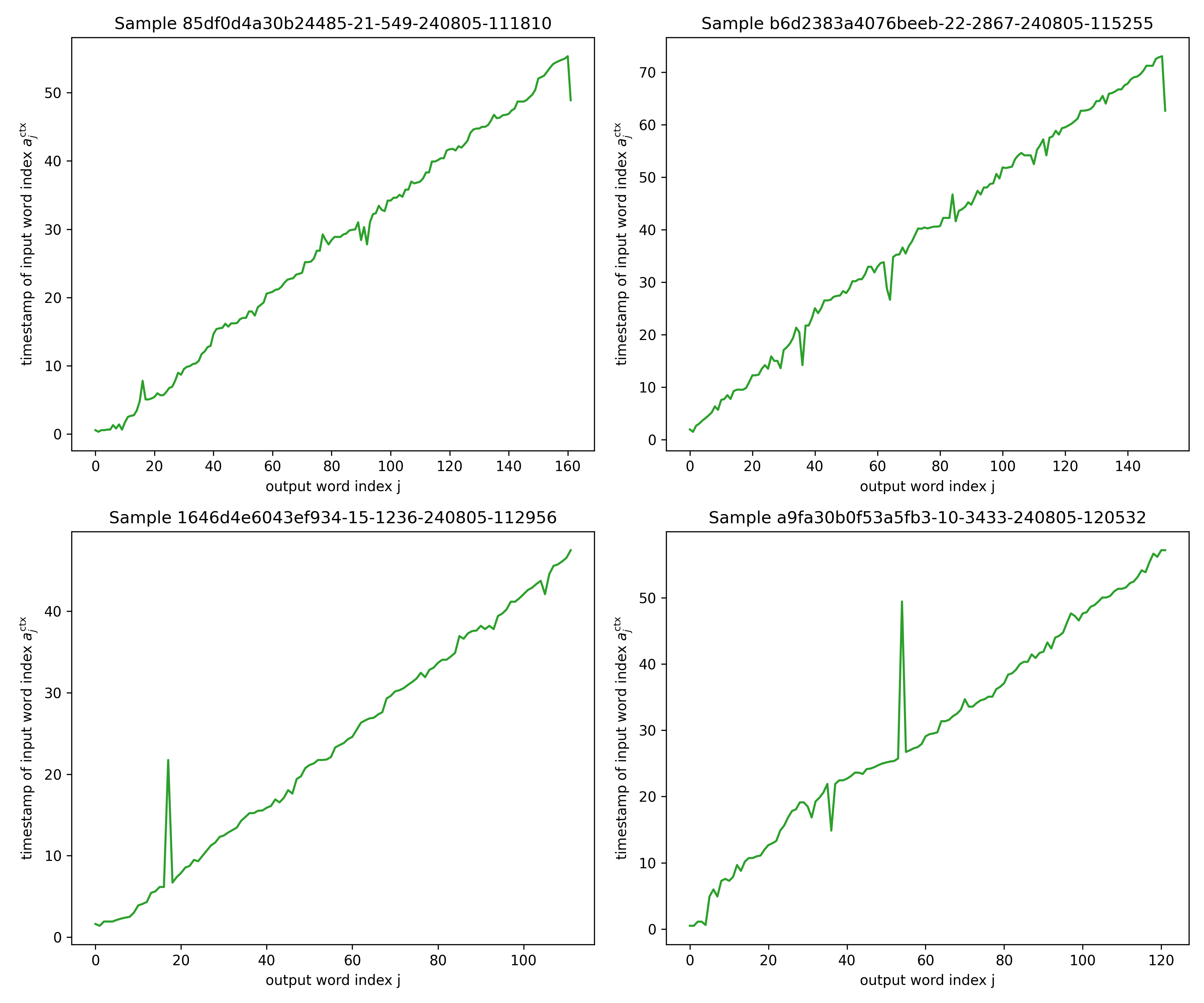}
    \caption{\textbf{Example of contextual alignments.} We compute the contextual alignment ($a_j^{\text{ctx}}$) for four different samples and plot the associated input timestamps. Some results as the two at the bottom present extreme spikes meaning output words referring to input words very far in the future. These spikes are considered as anomalies and are smoothed out as explained in Section \ref{sec:audio-domain-align}.}
    \label{fig:examples-contextual-alignments}
\end{figure}


\end{document}